\documentclass[11pt]{article}

\usepackage[final]{acl}

\usepackage{times}
\usepackage{latexsym}

\usepackage[T1]{fontenc}

\usepackage[utf8]{inputenc}

\usepackage{microtype}

\usepackage{inconsolata}

\usepackage{graphicx}

\usepackage{booktabs}
\usepackage{amsmath}
\usepackage{makecell}
\usepackage{amssymb}
\usepackage{multirow}
\usepackage{booktabs}
\usepackage{array}
\usepackage{enumitem}

\title{Confidence-Driven Multi-Scale Model Selection
for Cost-Efficient Inference}

\author{
    Bo-Wei Chen,\textsuperscript{\rm 1}
    Chung-Chi Chen,\textsuperscript{\rm 2}
    An-Zi Yen\textsuperscript{\rm 1}
    \\
    \textsuperscript{\rm 1} Department of Computer Science, National Yang Ming Chiao Tung University, Taiwan
    \\
    \textsuperscript{\rm 2} Artificial Intelligence Research Center, AIST, Japan
    \\
    \texttt{h7a4n1k.cs12@nycu.edu.tw,}
    \texttt{c.c.chen@acm.org,}
    \texttt{azyen@nycu.edu.tw}
}

\begin{document}
\maketitle
\begin{abstract}
Large Language Models (LLMs) have revolutionized inference across diverse natural language tasks, with larger models performing better but at higher computational costs.
We propose a confidence-driven strategy that dynamically selects the most suitable model based on confidence estimates.
By assessing a model's confidence in handling the task and response accuracy, tasks that are likely to be solved correctly are retained, while more uncertain or complex cases are delegated to a larger model, ensuring reliability while minimizing computation.
Specifically, we evaluate a model's likelihood of knowing the correct answer and the probability that its response is accurate.
Experiments on the Massive Multitask Language Understanding (MMLU) benchmark show that our approach achieves accuracy comparable to the largest model while reducing computational costs by 20\% to 40\%.
When applied to GPT-4o API calls, it reduces token usage by approximately 60\%, further improving cost efficiency. 
These findings indicate the potential of confidence-based model selection to enhance real-world LLM deployment, particularly in resource-constrained settings such as edge devices and commercial API applications.
\end{abstract}

\section{Introduction}

As large language models (LLMs) become increasingly integrated into real-world applications, their deployment poses significant challenges in terms of computational cost and efficiency \cite{bender2021dangers,chen2023frugalgptuselargelanguage}.
While offering significant capability, an LLM query often requires forwarding a 10 to 100-billion-scale model.
Moreover, with prompt engineering techniques such as in-context learning \cite{gpt3} or chain-of-thought reasoning \cite{wang2023selfconsistencyimproveschainthought}, reasoning tasks can demand additional tokens, resulting in inputs spanning thousands of tokens.
Therefore, running a large-scale model for every query is impractical, particularly for businesses, individual developers, and resource-constrained edge devices.
For example, on smartphones, local models offer faster response times due to reduced latency and smaller parameter sizes, but they often sacrifice accuracy and performance. 
Conversely, querying API with a more capable model (i.e., GPT-4o \cite{hurst2024gpt}) ensures higher-quality responses but introduces trade-offs such as network latency and increased computational expenses.
Given the variance in model capabilities and the differing complexity of tasks, selecting the most cost-effective model that can still deliver an adequate response has become a critical research problem. 
Prior approaches have explored selective routing of queries, but an optimal balance between efficiency and accuracy remains an open challenge.

\begin{figure*}[t]
  \centering
  \includegraphics[width=\textwidth]{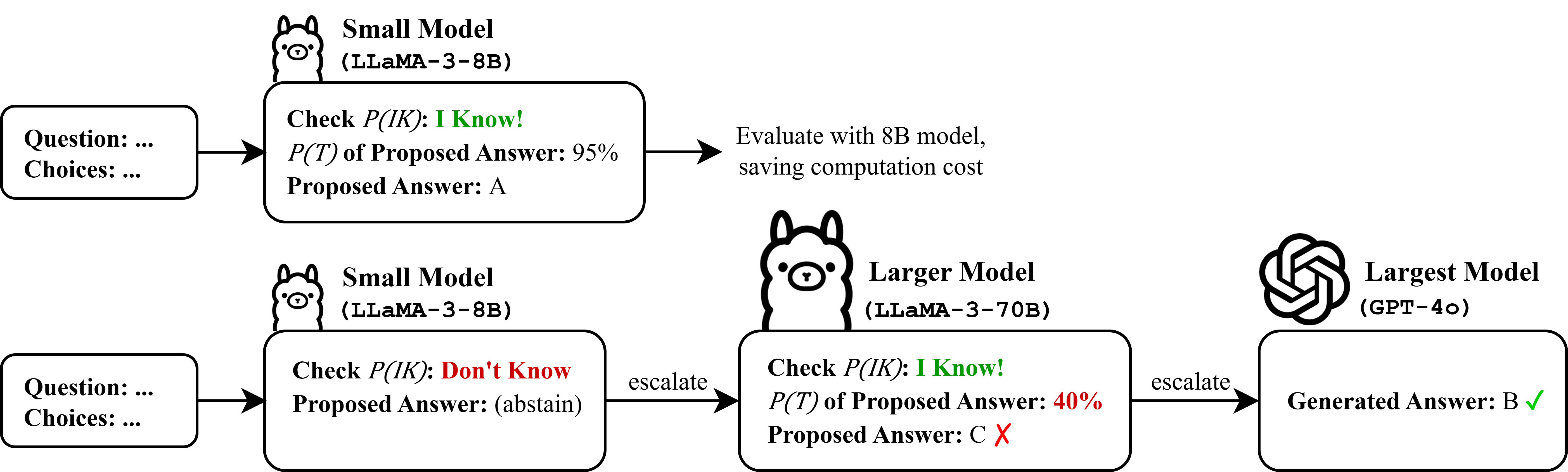}
  \caption{An illustration of our strategy.}
  \label{fig:example}
\end{figure*}

To address this, we propose a confidence-based model selection framework that optimizes efficiency while maintaining accuracy.
Our approach begins by querying the smallest LLM and escalates to larger models if the initial model lacks confidence in its prediction.
Inspired by \citet{kadavath2022languagemodelsmostlyknow}, we measure two facets of confidence: (1) $P(T)$ (Probability of Truth), which quantifies a model's confidence in the truthfulness of a given statement, and (2) $P(IK)$ (Probability of ``I Know''), which measures the likelihood that a model can correctly answer a given query, as shown in Figure~\ref{fig:example}.
By strategically orchestrating models of varying sizes, we achieve better cost-effectiveness while maintaining performance benchmarks on the Massive Multitask Language Understanding (MMLU) dataset \cite{hendrycks2021measuringmassivemultitasklanguage}.
We also analyze performance on an out-of-distribution (OOD) dataset to verify robustness.

In sum, our contributions are as follows. 
(1)~A confidence-driven model selection strategy is proposed to reduce computational cost while maintaining accuracy.\footnote{\url{https://github.com/NYCU-NLP-Lab/ConfDrivenInference}} 
(2)~When applied to commercial APIs such as GPT-4o, the method reduces token costs with minimal performance degradation. 
(3)~We demonstrate the generalizability of our approach to OOD datasets, showing its applicability beyond the training distribution.

\section{Methodology}

\label{ch:methodology}

The core challenge in effectively selecting LLMs to balance cost and capability lies in measuring the confidence of smaller LLMs and using it as the basis for escalation decisions.
Researchers \cite{kadavath2022languagemodelsmostlyknow, lin2022teachingmodelsexpressuncertainty,kuhn2023semanticuncertaintylinguisticinvariances, chen-mueller-2024-quantifying} have investigated whether a model can express precise confidence levels in their responses and assess their ability.
Previous work \cite{chen2023frugalgptuselargelanguage} uses LLM cascading with a trained scoring function based on model outputs, but neglects internal model information, such as token probabilities and hidden states, which could improve confidence estimation and decision making.
\cite{kadavath2022languagemodelsmostlyknow} show that language models produce well-calibrated probabilities when selecting among multiple explicit choices, with strong empirical results across benchmarks.
These results suggest that the confidence estimates of the model can serve as a reliable uncertainty signal.
Inspired by their work, we measure LLM confidences with $P(T)$ and $P(IK)$ of LLMs, where the former can be estimated without auxiliary models or additional training.\footnote{Further discussions on confidence measurement methods, routing, and hybrid inference are provided in Appendix~\ref{sec:related_work}.}

\subsection{Model Selection Process}
Suppose the user has access to $k$ LLMs, denoted as $M_1, M_2, \dots, M_k$, sorted in ascending order by the number of parameters, where the largest model is $M_k$. 
Instead of always using $M_k$'s output, our method first assesses $P(IK)$ for smaller models to determine whether they can confidently solve the query. 
Next, to verify the correctness of the response, we evaluate the model's $P(T)$ score to determine its confidence in the candidate answer. 
Specifically, if $P(IK)$ or $P(T)$ for $M_i$ fall below a predefined threshold, the query is passed to the next larger model, $M_{i+1}$.
Otherwise, $M_i$ is used to answer the query, thereby reducing the computational cost associated with larger models. 
This process continues until the query reaches the largest model ($M_k$), at which point the response from $M_k$ is used as the final answer.
Figure \ref{fig:example} shows our strategy of escalating uncertain queries to larger models to improve efficiency.

\subsection{Confidence Measurement} 
To determine $P(T)$, we measure the probability of the first token after prompting the model with a question and answer choices. 
Given a formatted prompt $x$, $P(T)$ is the probability that the predicted token belongs to the answer set. 
The final answer is determined based on the highest \textbf{$P(T)$}:
\begin{equation}
    P(T) = p(c \mid x), \quad c \in \{A, B, C, D\}.
\end{equation}

The value of $P(T)$ is directly obtained from the model's probability distribution over answer choices, corresponding to the predicted probability of the most confident option.
For generation tasks, we use a different prompt that asks the model to answer the query in a predefined format \textit{``Answer: ''}, and measure the probability of the first token being \textit{``Answer''}.
To establish a threshold for $P(T)$, we conducted sensitivity analysis to identify suitable values. 
Based on the observations from the analysis, we set the threshold to 0.9 for all experiments in this work.

To compute $P(IK)$, we trained a multi-layer perceptron using hidden states from the 24th transformer layer of an LLM as input \cite{mahaut-etal-2024-factual}, with supervision indicating whether the model correctly answered each query.
The model is trained on 80\% of the dataset, with 10\% used for validation and 10\% for testing across all datasets in this work, unless otherwise specified.

\section{Experiments}
\label{sec:experiments}

\subsection{Experimental Setup}
We evaluate our strategy on MMLU, a multiple-choice benchmark assessing general knowledge, reasoning, and problem-solving across 57 subjects in humanities, sciences, and other domains.

For reproducibility, we conduct experiments using the  open-source LLaMA instruction-tuned models \cite{grattafiori2024llama3herdmodels}, and Qwen 3 series \cite{yang2025qwen3technicalreport}, which are available in sizes of (3B, 8B, 70B) and (4B, 8B, 32B) respectively.
The specific open-source models used in our experiments are as follows: Llama-3.2-3B-Instruct, Meta-Llama-3.1-8B-Instruct, Llama-3.3-70B-Instruct, Qwen3-4B, Qwen3-8B, Qwen3-32B.\footnote{Model details are presented in Appendix~\ref{sec:models}.}
Additionally, we incorporate GPT-4o to explore its applicability in commercial API scenarios.\footnote{Prompts used in this work are provided in Appendix~\ref{sec:prompt}.}

Accuracy is measured as the proportion of questions for which the predicted answer matches the ground truth.
In addition to accuracy, we assess the cost-effectiveness of different model combinations by measuring the end-to-end execution time, and calculating the computational cost of evaluating the models.
To estimate the computational cost of the forward pass in LLMs, we adopt \cite{kaplan2020scalinglawsneurallanguage}'s approach.
The number of add-multiply operations required for a forward pass in a Transformer-based model is given by:
\begin{equation}
C_{\text{forward}} \approx 2N + 2n_{\text{layer}} \times  n_{\text{ctx}} \times d_{\text{model}}\,,
\end{equation}
where $N$ is the number of model parameters, $n_{\text{ctx}}$ is the number of context tokens, and $d_{\text{model}}$ denotes the dimension of the residual stream. For simplification, we approximate $N$ using:
\begin{equation}
N \approx 12 \times n_{\text{layer}} \times  d_{\text{model}}^2\,,
\end{equation}

Substituting into the original equation gives: 
\begin{equation}
C_{\text{forward}} \approx 2N + 2n_{\text{ctx}} \times  \sqrt{\frac{N n_{\text{layer}}}{12}}\,,
\end{equation}

For large values of $N$ (as model sizes are typically in the billions of parameters), the context-dependent term becomes significant only when $n_{\text{ctx}}$ is comparable to $N$. 
Thus, the complexity of a forward pass is approximately: 
\begin{equation}
C_{\text{forward}} \approx 2N
\end{equation}

To clarify the differences in computational cost (CC) across model configurations, we provide the specific number of queries handled by each model stage during evaluation. The compute cost is defined as:
\begin{equation}
\text{CC} = 2 \times N \times k
\end{equation}
where $N$ is the number of model parameters and $k$ is the number of queries evaluated by that model.
The factor of 2 accounts for the two operations (multiply and accumulate) in the matrix multiplications. 
Appendix~\ref{sec:q_ct} presents the number of queries across different  stages.

\subsection{Results with Multi-Scale Open-Source Models}

\begin{table}[t]
  \centering
  \resizebox{\columnwidth}{!}{
    \begin{tabular}{lrccc}
    \toprule
    \multicolumn{1}{c}{\textbf{LLM(s)}} & \multicolumn{1}{c}{\textbf{Acc.}} & \textbf{PD ($\downarrow$)} & \textbf{Reduced CC ($\uparrow$)} & \textbf{Time ($\downarrow$)}\\
    \hline
    3B    & 63.78\% & -- & -- & -- \\
    8B    & 70.21\% & -- & -- & -- \\
    70B   & 83.57\% & -- & -- & -- \\
    \hline
    3B $\to$ 8B & 69.44\% & -0.77\% & 20.34\% & -18.90\%  \\
    3B $\to$ 70B & 81.89\% & -1.68\% & 32.47\% & -26.51\%  \\
    8B $\to$ 70B & 83.22\% & -0.35\% & 36.46\% & -33.03\%  \\
    \hline
    3B $\to$ 8B $\to$ 70B & 81.54\% & -2.03\% & 44.48\% & -- \\
    \bottomrule
    \end{tabular}
  }
  \caption{Experimental results with LLaMA models on MMLU. Acc., Reduced CC, Time, and PD denote accuracy, reduced computational cost (measured in GFLOPs), end-to-end time, and performance drop. Each metric is calculated relative to those of $M_k$.}
  \label{tab:main-exp}
\end{table}

Table~\ref{tab:main-exp} presents the experimental results using LLaMA models.
To provide a more comprehensive assessment beyond accuracy, we also report Macro-Precision, Macro-Recall, and Macro-F1 scores for all model configurations. 
These metrics offer a class-agnostic view of performance by treating each class equally, regardless of its frequency.
The results are presented in Table~\ref{tab:macro-metrics}.\footnote{The end-to-end time for the 3B $\to$ 8B $\to$ 70B chain is omitted due to computation resource constraints. }

\begin{table}[t]
  \centering
  \small
    \begin{tabular}{lccc}
    \toprule
    \textbf{LLM(s)} & \textbf{P} & \textbf{R} & \textbf{F1} \\
    \hline
    3B & 0.6444 & 0.6372 & 0.6368 \\
    8B & 0.7055 & 0.7021 & 0.7017 \\
    70B & 0.8386 & 0.8354 & 0.8357 \\
    \hline
    3B $\to$ 8B & 0.6974 & 0.6941 & 0.6940 \\
    3B $\to$ 70B & 0.8198 & 0.8160 & 0.8166 \\
    8B $\to$ 70B & 0.8350 & 0.8317 & 0.8322 \\
    \hline
    3B $\to$ 8B $\to$ 70B & 0.8162 & 0.8124 & 0.8130 \\
    \bottomrule
    \end{tabular}
\caption{Macro-Precision (P), Macro-Recall (R), and Macro-F1 (F1) scores of LLaMA models on MMLU.}
\label{tab:macro-metrics}
\end{table}

Several key observations can be made.
First, as expected, utilizing the largest 70B model yields the best performance. 
However, this comes at a significantly higher computational cost compared to smaller models, highlighting the inherent trade-off between performance and efficiency. 
Second, by employing the proposed confidence-driven strategy, computational complexity is notably reduced with only a minor drop in performance.
A similar reduction is also observed in the end-to-end execution time.
This demonstrates the potential of leveraging LLM confidence for multi-scale model selection.
Third, comparing the results of (8B, 3B $\to$ 8B) and (8B, 8B $\to$ 70B), we observe that applying the proposed strategy with a model one size smaller results in minimal performance degradation while achieving an approximate 20-36\% reduction in computational cost. 
Notably, 8B $\to$ 70B achieves comparable accuracy to 70B (83.22\% vs. 83.57\%) with a 36\% reduction in cost.
We calculate McNemar's statistical significance test and observe no significant difference ($p = 0.4048$), supporting the efficiency of our approach.\footnote{Appendix~\ref{sec:p_value} presents detailed significance test results.}
However, extending the transfer chain (8B $\to$ 70B and 3B $\to$ 8B $\to$ 70B) leads to a more pronounced performance drop, but with increased cost savings. 
This is expected, as errors may accumulate and propagate under such a design.  
Additionally, in the 3B $\to$ 8B $\to$ 70B process, the smaller model (e.g., 3B) may overestimate its confidence, leading to premature task retention and suboptimal delegation, ultimately affecting the final reasoning outcome.

\begin{table}[t]
  \centering
  \resizebox{\columnwidth}{!}{
    \begin{tabular}{lrccc}
    \toprule
    \multicolumn{1}{c}{\textbf{LLM(s)}} & \multicolumn{1}{c}{\textbf{Acc.}} & \textbf{PD ($\downarrow$)} & \textbf{Reduced CC ($\uparrow$)} & \textbf{Time ($\downarrow$)} \\
    \hline
    4B    & 66.73\% & -- & -- & -- \\
    8B    & 72.43\% & -- & -- & -- \\
    32B   & 79.51\% & -- & -- & -- \\
    \hline
    4B $\to$ 8B & 70.21\% & -3.07\% & 17.80\% & 2.55\% \\
    4B $\to$ 32B & 75.03\% & -5.63\% & 52.70\% & -40.21\% \\
    8B $\to$ 32B & 80.00\% & 0.62\% & 33.18\% & -23.76\% \\
    \hline
    4B $\to$ 8B $\to$ 32B & 74.69\% & -6.06\% & 55.06\% & -- \\
    \bottomrule
    \end{tabular}
    }
  \caption{Experimental results with Qwen 3 models on MMLU.}
  \label{tab:main-exp-qwen}
\end{table}

As shown in the Table~\ref{tab:main-exp-qwen}, our proposed confidence-based model selection method also demonstrates promising results on the Qwen 3 series.\footnote{The end-to-end time for the 4B $\to$ 8B $\to$ 32B chain is omitted due to computation resource constraints. }
Specifically, the 8B $\to$ 32B cascade achieves an accuracy of 0.800, which is slightly higher than the full 32B model (0.795), while saving approximately 33.18\% of the compute. This confirms the generalizability of our method beyond the LLaMA series and its effectiveness across different model families.
We further extend our method to open-ended question answering, demonstrating promising results on PopQA with a 7\% reduction in cost and less than a 2\% drop in performance.
The detailed results are provided in Appendix~\ref{sec:open-end}.
To sum up, these findings support the effectiveness of the confidence-driven strategy in balancing performance and computational cost.

\subsection{Results with Commercial API}
To evaluate our strategy in a more realistic scenario, we tested it using a commercial API. 
Since OpenAI does not disclose the total parameter count for GPT-4o, we use the number of output tokens as a proxy for the compute cost metric. 
This metric aligns with mainstream API cost calculation methods. 

Table \ref{tab:exp-api} compares the results of using GPT-4o as the final model in our method versus directly prompting GPT-4o. 
While other approaches result in a slight performance drop for a 15--20\% cost reduction, the 70B $\to$ GPT-4o approach slightly improves performance and achieves approximately 60\% cost savings.
Our strategy effectively balances performance and cost, particularly by leveraging an intermediate 70B model before GPT-4o, achieving significant token cost reduction while maintaining comparable performance. Even alternative approaches with minor performance loss provide meaningful savings, demonstrating the practicality of our method for optimizing API usage.\footnote{Appendix~\ref{sec:cost} provides an evaluation of cost across settings.}

\begin{table}[t]
  \centering
  \scriptsize
  \setlength\tabcolsep{2mm}
    \begin{tabular}{lrrcc}
    \toprule
    \multicolumn{1}{c}{\textbf{LLM(s)}} & \multicolumn{1}{c}{\textbf{Acc.}} &  \multicolumn{1}{c}{\textbf{Tokens}} & \textbf{PD} & \textbf{Reduced Tokens} \\
    \hline
    GPT-4o & 86.43\% & 36,225 & \multicolumn{1}{c}{--} & \multicolumn{1}{c}{--} \\
    \hline
    3B $\to$ GPT-4o & 84.48\% & 30,671 & -1.95\% & 15.33\% \\
    8B $\to$ GPT-4o & 85.66\% & 28,941 & -0.77\% & 20.11\% \\
    70B $\to$ GPT-4o & 86.85\% & 14,505 & 0.42\% & 59.96\% \\
    \bottomrule
    \end{tabular}%
  \caption{Experimental results with GPT-4o.}
  \label{tab:exp-api}%
\end{table}%

\section{Discussion}
\label{ch:discussion}

\subsection{Ablation Analysis}

In this section, we present an ablation study on $P(IK)$, focusing on performance drop since the transfer chain cost remains similar with or without $P(IK)$.\footnote{Results of $P(IK)$ are reported in Appendix~\ref{sec:result_pik}.}
Table~\ref{tab:Ablation} presents the experimental results.
First, the results indicate the usefulness of training a classifier to assess $P(IK)$. 
Specifically, the performance drop is significantly greater when using the transfer chain with the 70B model as the largest model without $P(IK)$.
Second, even without $P(IK)$, which requires training data for assessment, the performance drop remains smaller than the computational cost saved. 
These results validate the effectiveness of the confidence-driven multi-scale model selection, even without training data.
This finding also highlights the method's practicality in closed-source or API-based settings, where internal hidden states are not accessible and $P(IK)$ cannot be computed.
In such cases, $P(T)$ can still be derived from log probabilities provided by the model.
As shown in Table~\ref{tab:Ablation}, relying solely on $P(T)$ still achieves competitive performance.\footnote{Results for different $P(T)$ thresholds are in Appendix~\ref{sec:sensitivity}.}

\begin{table}[t]
  \centering
  \small
  \setlength{\tabcolsep}{1mm}
    \begin{tabular}{llcc}
    \toprule
    \textbf{Setting} & \multicolumn{1}{c}{\textbf{LLM(s)}} & \textbf{PD} & \textbf{Reduced CC} \\
    \hline
    \multirow{2}[1]{*}{w/ $P(IK)$} & 8B $\to$ 70B & -0.35\% & 36.46\% \\
          & 3B $\to$ 8B $\to$ 70B & -2.03\% & 44.48\% \\
    \hline
    \multirow{2}[1]{*}{w/o $P(IK)$} & 8B $\to$ 70B & -1.26\% & 36.83\% \\
          & 3B $\to$ 8B $\to$ 70B & -4.27\% & 47.17\% \\
    \bottomrule
    \end{tabular}%
  \caption{Ablation studies of $P(IK)$.}
  \label{tab:Ablation}%
\end{table}%

\subsection{Generalization to OOD Dataset} 
A key question when evaluating a $P(IK)$ classifier trained on MMLU is whether it can generalize to other datasets.
Hence, we investigate if a classifier that assesses a model's NLU across different question types can be directly applied to new datasets without modification.
Such a finding would suggest that a well-trained $P(IK)$ classifier captures a model's intrinsic ability for diverse NLU tasks, independent of the specific dataset characteristics.

To explore this hypothesis, we employ GPQA: A Graduate-Level Google-Proof Q\&A Benchmark~\cite{rein2023gpqagraduatelevelgoogleproofqa}. 
GPQA represents an out-of-distribution (OOD) setting relative to MMLU, as it comprises highly challenging, graduate-level questions. 
By applying the same $P(IK)$ classification approach to GPQA, we aim to assess whether the classifier retains its effectiveness in measuring model performance in an OOD context.

\begin{table}[t]
  \centering
  \small
    \begin{tabular}{lccc}
    \toprule
    \multicolumn{1}{c}{\textbf{LLM(s)}} & \textbf{Acc.} & \textbf{PD} & \textbf{Reduced CC} \\
    \hline
    3B    & 31.47\% & -- & -- \\
    8B    & 31.92\% & -- & -- \\
    70B   & 52.23\% & -- & -- \\
    \hline
    8B $\to$ 70B & 51.79\% & -0.44\% & 3.93\% \\
    3B $\to$ 8B $\to$ 70B & 51.56\% & -0.67\% & 5.11\% \\
    \bottomrule
    \end{tabular}
  \caption{Experimental results of LLaMA models in OOD scenario.}
  \label{tab:OOD}
\end{table}

Table~\ref{tab:OOD} presents the experimental results. 
First, GPQA proves to be significantly more challenging than MMLU for all model scales. 
Second, the proposed framework results in a modest decrease in computational costs while maintaining competitive performance.
However, the substantial difference in computational cost savings between MMLU and GPQA highlights the difficulty of extending the method to OOD applications.
This may be due to the fact that GPQA's questions are too challenging for smaller models, such as 3B and 8B, making it difficult for them to effectively handle queries. 
We looked into the amount of queries that the $P(IK)$ classifier predicts positive in the 8B $\to$ 70B case.
Out of 448 samples in the test set, only 21 samples were routed to 8B by the classifier, showing that the classifier is conservative while facing OOD queries rather than aggressively routing to the 8B model.
However, among these 21 samples, 67\% of them (14 samples) are ``correct'', meaning that the 8B model can indeed handle these queries, showing that the $P(IK)$ classifier remains useful in the OOD scenario.
This preliminary exploration identifies this challenge for future research in advancing confidence-driven multi-scale model selection.

\section{Conclusions}
\label{ch:conclusion}
We propose a confidence-driven multi-scale model selection strategy to optimize cost-effective inference.
By combining confidence estimates of $P(T)$ (token probabilities) and $P(IK)$ (trained classifiers), our approach dynamically routes queries to the most appropriate model, significantly reducing computational costs while maintaining high accuracy.
Our experiments evaluating LLaMA 3 and Qwen 3 models on MMLU show that the proposed strategy matches the performance of the largest model while reducing compute cost.
When applied to GPT-4o, it significantly cuts token usage and improves cost efficiency.
Moreover, we demonstrate that our approach remains effective in both in-distribution and out-of-distribution settings by leveraging classifiers trained with MMLU on GPQA, a significantly more challenging dataset.
Future work should focus on refining confidence estimation for better generalization across diverse tasks, particularly generative tasks.

\section*{Limitations}
\label{ch:limitation}

First, our approach relies on a classifier trained on a specific dataset to estimate $P(IK)$. While we observe some level of generalization to out-of-distribution datasets such as GPQA, performance may degrade in significantly different domains or tasks. Future work should explore methods to improve generalization, such as domain adaptation techniques or unsupervised confidence estimation.
Second, our experiments primarily focus on NLU tasks within the MMLU and GPQA benchmarks.
Our experiments with PopQA served as a preliminary attempt to generalize confidence-based routing beyond NLU tasks.
The applicability of our method to other NLP tasks, such as generative language modeling, open-domain QA, or real-time conversational AI, remains an open challenge.
Future work should investigate extending confidence-based model selection to broader NLP applications.
Third, while our method reduces overall compute usage, it may introduce additional latency for queries that are escalated through multiple models before reaching the final response. Although we use compute cost as a proxy for efficiency, actual end-to-end latency can be affected by system implementation, hardware, and network conditions, especially in API-based settings. Future work could incorporate latency-aware routing or empirically evaluate real-world latency across deployment scenarios.
Despite these limitations, our findings indicate that confidence-driven model selection is a promising direction for optimizing LLM deployment, particularly in scenarios where computational efficiency is a critical concern. Addressing these challenges will be key to further improving the reliability and applicability of the approach across diverse settings.

\section*{Acknowledgments}
This research was partially supported by the National Science and Technology Council, Taiwan, under Grant NSTC 114-2221-E-A49-057-MY3, and by the project ``Advancements in Industrial and Financial Large Language Models: Construction and Applications,'' funded by the Research Center for Information Technology Innovation (CITI), Academia Sinica.
Chung-Chi Chen's work was supported by AIST policy-based budget project ``R\&D on Generative AI Foundation Models for the Physical Domain''.

\bibliography{custom}

@misc{kadavath2022languagemodelsmostlyknow,
      title={Language Models (Mostly) Know What They Know}, 
      author={Saurav Kadavath and Tom Conerly and Amanda Askell and Tom Henighan and Dawn Drain and Ethan Perez and Nicholas Schiefer and Zac Hatfield-Dodds and Nova DasSarma and Eli Tran-Johnson and Scott Johnston and Sheer El-Showk and Andy Jones and Nelson Elhage and Tristan Hume and Anna Chen and Yuntao Bai and Sam Bowman and Stanislav Fort and Deep Ganguli and Danny Hernandez and Josh Jacobson and Jackson Kernion and Shauna Kravec and Liane Lovitt and Kamal Ndousse and Catherine Olsson and Sam Ringer and Dario Amodei and Tom Brown and Jack Clark and Nicholas Joseph and Ben Mann and Sam McCandlish and Chris Olah and Jared Kaplan},
      year={2022},
      eprint={2207.05221},
      archivePrefix={arXiv},
      primaryClass={cs.CL},
      url={https://arxiv.org/abs/2207.05221}, 
}

@misc{xiong2024llmsexpressuncertaintyempirical,
      title={Can LLMs Express Their Uncertainty? An Empirical Evaluation of Confidence Elicitation in LLMs}, 
      author={Miao Xiong and Zhiyuan Hu and Xinyang Lu and Yifei Li and Jie Fu and Junxian He and Bryan Hooi},
      year={2024},
      eprint={2306.13063},
      archivePrefix={arXiv},
      primaryClass={cs.CL},
      url={https://arxiv.org/abs/2306.13063}, 
}

@inproceedings{mahaut-etal-2024-factual,
    title = "Factual Confidence of {LLM}s: on Reliability and Robustness of Current Estimators",
    author = {Mahaut, Mat{\'e}o  and
      Aina, Laura  and
      Czarnowska, Paula  and
      Hardalov, Momchil  and
      M{\"u}ller, Thomas  and
      Marquez, Lluis},
    editor = "Ku, Lun-Wei  and
      Martins, Andre  and
      Srikumar, Vivek",
    booktitle = "Proceedings of the 62nd Annual Meeting of the Association for Computational Linguistics (Volume 1: Long Papers)",
    month = aug,
    year = "2024",
    address = "Bangkok, Thailand",
    publisher = "Association for Computational Linguistics",
    url = "https://aclanthology.org/2024.acl-long.250/",
    doi = "10.18653/v1/2024.acl-long.250",
    pages = "4554--4570"
}

@misc{lin2022teachingmodelsexpressuncertainty,
      title={Teaching Models to Express Their Uncertainty in Words}, 
      author={Stephanie Lin and Jacob Hilton and Owain Evans},
      year={2022},
      eprint={2205.14334},
      archivePrefix={arXiv},
      primaryClass={cs.CL},
      url={https://arxiv.org/abs/2205.14334}, 
}

@inproceedings{chen-mueller-2024-quantifying,
    title = "Quantifying Uncertainty in Answers from any Language Model and Enhancing their Trustworthiness",
    author = "Chen, Jiuhai  and
      Mueller, Jonas",
    editor = "Ku, Lun-Wei  and
      Martins, Andre  and
      Srikumar, Vivek",
    booktitle = "Proceedings of the 62nd Annual Meeting of the Association for Computational Linguistics (Volume 1: Long Papers)",
    month = aug,
    year = "2024",
    address = "Bangkok, Thailand",
    publisher = "Association for Computational Linguistics",
    url = "https://aclanthology.org/2024.acl-long.283/",
    doi = "10.18653/v1/2024.acl-long.283",
    pages = "5186--5200"
}

@misc{rein2023gpqagraduatelevelgoogleproofqa,
      title={GPQA: A Graduate-Level Google-Proof Q\&A Benchmark}, 
      author={David Rein and Betty Li Hou and Asa Cooper Stickland and Jackson Petty and Richard Yuanzhe Pang and Julien Dirani and Julian Michael and Samuel R. Bowman},
      year={2023},
      eprint={2311.12022},
      archivePrefix={arXiv},
      primaryClass={cs.AI},
      url={https://arxiv.org/abs/2311.12022}, 
}

@misc{ren2023selfevaluationimprovesselectivegeneration,
      title={Self-Evaluation Improves Selective Generation in Large Language Models}, 
      author={Jie Ren and Yao Zhao and Tu Vu and Peter J. Liu and Balaji Lakshminarayanan},
      year={2023},
      eprint={2312.09300},
      archivePrefix={arXiv},
      primaryClass={cs.CL},
      url={https://arxiv.org/abs/2312.09300}, 
}

@inproceedings{tian-etal-2023-just,
    title = "Just Ask for Calibration: Strategies for Eliciting Calibrated Confidence Scores from Language Models Fine-Tuned with Human Feedback",
    author = "Tian, Katherine  and
      Mitchell, Eric  and
      Zhou, Allan  and
      Sharma, Archit  and
      Rafailov, Rafael  and
      Yao, Huaxiu  and
      Finn, Chelsea  and
      Manning, Christopher",
    editor = "Bouamor, Houda  and
      Pino, Juan  and
      Bali, Kalika",
    booktitle = "Proceedings of the 2023 Conference on Empirical Methods in Natural Language Processing",
    month = dec,
    year = "2023",
    address = "Singapore",
    publisher = "Association for Computational Linguistics",
    url = "https://aclanthology.org/2023.emnlp-main.330/",
    doi = "10.18653/v1/2023.emnlp-main.330",
    pages = "5433--5442"
}

@misc{wang2023selfconsistencyimproveschainthought,
      title={Self-Consistency Improves Chain of Thought Reasoning in Language Models}, 
      author={Xuezhi Wang and Jason Wei and Dale Schuurmans and Quoc Le and Ed Chi and Sharan Narang and Aakanksha Chowdhery and Denny Zhou},
      year={2023},
      eprint={2203.11171},
      archivePrefix={arXiv},
      primaryClass={cs.CL},
      url={https://arxiv.org/abs/2203.11171}, 
}

@misc{kuhn2023semanticuncertaintylinguisticinvariances,
      title={Semantic Uncertainty: Linguistic Invariances for Uncertainty Estimation in Natural Language Generation}, 
      author={Lorenz Kuhn and Yarin Gal and Sebastian Farquhar},
      year={2023},
      eprint={2302.09664},
      archivePrefix={arXiv},
      primaryClass={cs.CL},
      url={https://arxiv.org/abs/2302.09664}, 
}

@article{grattafiori2024llama3herdmodels,
  title={The llama 3 herd of models},
  author={Dubey, Abhimanyu and Jauhri, Abhinav and Pandey, Abhinav and Kadian, Abhishek and Al-Dahle, Ahmad and Letman, Aiesha and Mathur, Akhil and Schelten, Alan and Yang, Amy and Fan, Angela and others},
  journal={arXiv preprint arXiv:2407.21783},
  url={https://arxiv.org/abs/2407.21783}, 
  year={2024}
}

@misc{kaplan2020scalinglawsneurallanguage,
      title={Scaling Laws for Neural Language Models}, 
      author={Jared Kaplan and Sam McCandlish and Tom Henighan and Tom B. Brown and Benjamin Chess and Rewon Child and Scott Gray and Alec Radford and Jeffrey Wu and Dario Amodei},
      year={2020},
      eprint={2001.08361},
      archivePrefix={arXiv},
      primaryClass={cs.LG},
      url={https://arxiv.org/abs/2001.08361}, 
}

@misc{hendrycks2021measuringmassivemultitasklanguage,
      title={Measuring Massive Multitask Language Understanding}, 
      author={Dan Hendrycks and Collin Burns and Steven Basart and Andy Zou and Mantas Mazeika and Dawn Song and Jacob Steinhardt},
      year={2021},
      eprint={2009.03300},
      archivePrefix={arXiv},
      primaryClass={cs.CY},
      url={https://arxiv.org/abs/2009.03300}, 
}

@misc{chen2023frugalgptuselargelanguage,
      title={FrugalGPT: How to Use Large Language Models While Reducing Cost and Improving Performance}, 
      author={Lingjiao Chen and Matei Zaharia and James Zou},
      year={2023},
      eprint={2305.05176},
      archivePrefix={arXiv},
      primaryClass={cs.LG},
      url={https://arxiv.org/abs/2305.05176}, 
}

@inproceedings{bender2021dangers,
  title={On the dangers of stochastic parrots: Can language models be too big?},
  author={Bender, Emily M and Gebru, Timnit and McMillan-Major, Angelina and Shmitchell, Shmargaret},
  booktitle={Proceedings of the 2021 ACM conference on fairness, accountability, and transparency},
  pages={610--623},
  year={2021}
}

@article{hurst2024gpt,
  title={Gpt-4o system card},
  author={Hurst, Aaron and Lerer, Adam and Goucher, Adam P and Perelman, Adam and Ramesh, Aditya and Clark, Aidan and Ostrow, AJ and Welihinda, Akila and Hayes, Alan and Radford, Alec and others},
  journal={arXiv preprint arXiv:2410.21276},
  year={2024}
}

@misc{gpt3,
      title={Language Models are Few-Shot Learners}, 
      author={Tom B. Brown and Benjamin Mann and Nick Ryder and Melanie Subbiah and Jared Kaplan and Prafulla Dhariwal and Arvind Neelakantan and Pranav Shyam and Girish Sastry and Amanda Askell and Sandhini Agarwal and Ariel Herbert-Voss and Gretchen Krueger and Tom Henighan and Rewon Child and Aditya Ramesh and Daniel M. Ziegler and Jeffrey Wu and Clemens Winter and Christopher Hesse and Mark Chen and Eric Sigler and Mateusz Litwin and Scott Gray and Benjamin Chess and Jack Clark and Christopher Berner and Sam McCandlish and Alec Radford and Ilya Sutskever and Dario Amodei},
      year={2020},
      eprint={2005.14165},
      archivePrefix={arXiv},
      primaryClass={cs.CL},
      url={https://arxiv.org/abs/2005.14165}, 
}

@article{zhao2024eagle,
  title={Eagle: Efficient training-free router for multi-llm inference},
  author={Zhao, Zesen and Jin, Shuowei and Mao, Z Morley},
  journal={arXiv preprint arXiv:2409.15518},
  year={2024}
}

@inproceedings{ong2024routellm,
  title={RouteLLM: Learning to Route LLMs from Preference Data},
  author={Ong, Isaac and Almahairi, Amjad and Wu, Vincent and Chiang, Wei-Lin and Wu, Tianhao and Gonzalez, Joseph E and Kadous, M Waleed and Stoica, Ion},
  booktitle={The Thirteenth International Conference on Learning Representations},
  year={2024}
}

@article{oh2024uncertainty,
  title={Uncertainty-Aware Hybrid Inference with On-Device Small and Remote Large Language Models},
  author={Oh, Seungeun and Kim, Jinhyuk and Park, Jihong and Ko, Seung-Woo and Quek, Tony QS and Kim, Seong-Lyun},
  journal={arXiv preprint arXiv:2412.12687},
  year={2024}
}

@article{chen2025symbolic,
        title={Symbolic Mixture-of-Experts: Adaptive Skill-based Routing for Scalable Heterogeneous Reasoning},
        author={Chen, Justin Chih-Yao and Yun, Sukwon and Stengel-Eskin, Elias and Chen, Tianlong and Bansal, Mohit},
        journal={arXiv preprint arXiv:2503.05641},
        year={2025}}

@inproceedings{
shnitzer2024large,
title={Large Language Model Routing with Benchmark Datasets},
author={Tal Shnitzer and Anthony Ou and M{\'\i}rian Silva and Kate Soule and Yuekai Sun and Justin Solomon and Neil Thompson and Mikhail Yurochkin},
booktitle={First Conference on Language Modeling},
year={2024}
}

@inproceedings{dinghybrid,
  title={Hybrid LLM: Cost-Efficient and Quality-Aware Query Routing},
  author={Ding, Dujian and Mallick, Ankur and Wang, Chi and Sim, Robert and Mukherjee, Subhabrata and R{\"u}hle, Victor and Lakshmanan, Laks VS and Awadallah, Ahmed Hassan},
  year = {2024},
  booktitle={The Twelfth International Conference on Learning Representations}
}

@misc{yang2025qwen3technicalreport,
      title={Qwen3 Technical Report}, 
      author={An Yang and Anfeng Li and Baosong Yang and Beichen Zhang and Binyuan Hui and Bo Zheng and Bowen Yu and Chang Gao and Chengen Huang and Chenxu Lv and Chujie Zheng and Dayiheng Liu and Fan Zhou and Fei Huang and Feng Hu and Hao Ge and Haoran Wei and Huan Lin and Jialong Tang and Jian Yang and Jianhong Tu and Jianwei Zhang and Jianxin Yang and Jiaxi Yang and Jing Zhou and Jingren Zhou and Junyang Lin and Kai Dang and Keqin Bao and Kexin Yang and Le Yu and Lianghao Deng and Mei Li and Mingfeng Xue and Mingze Li and Pei Zhang and Peng Wang and Qin Zhu and Rui Men and Ruize Gao and Shixuan Liu and Shuang Luo and Tianhao Li and Tianyi Tang and Wenbiao Yin and Xingzhang Ren and Xinyu Wang and Xinyu Zhang and Xuancheng Ren and Yang Fan and Yang Su and Yichang Zhang and Yinger Zhang and Yu Wan and Yuqiong Liu and Zekun Wang and Zeyu Cui and Zhenru Zhang and Zhipeng Zhou and Zihan Qiu},
      year={2025},
      eprint={2505.09388},
      archivePrefix={arXiv},
      primaryClass={cs.CL},
      url={https://arxiv.org/abs/2505.09388}, 
}

@misc{chuang2025confidentseekstrongerexploring,
      title={Confident or Seek Stronger: Exploring Uncertainty-Based On-device LLM Routing From Benchmarking to Generalization}, 
      author={Yu-Neng Chuang and Leisheng Yu and Guanchu Wang and Lizhe Zhang and Zirui Liu and Xuanting Cai and Yang Sui and Vladimir Braverman and Xia Hu},
      year={2025},
      eprint={2502.04428},
      archivePrefix={arXiv},
      primaryClass={cs.CL},
      url={https://arxiv.org/abs/2502.04428}, 
}

@inproceedings{kamalloo-etal-2023-evaluating,
    title = "Evaluating Open-Domain Question Answering in the Era of Large Language Models",
    author = "Kamalloo, Ehsan  and
      Dziri, Nouha  and
      Clarke, Charles  and
      Rafiei, Davood",
    editor = "Rogers, Anna  and
      Boyd-Graber, Jordan  and
      Okazaki, Naoaki",
    booktitle = "Proceedings of the 61st Annual Meeting of the Association for Computational Linguistics (Volume 1: Long Papers)",
    month = jul,
    year = "2023",
    address = "Toronto, Canada",
    publisher = "Association for Computational Linguistics",
    url = "https://aclanthology.org/2023.acl-long.307/",
    doi = "10.18653/v1/2023.acl-long.307",
    pages = "5591--5606",
}

\appendix

\section{Related Work}\label{sec:related_work}

\subsection{Confidence Measurement Methods}
\citet{kadavath2022languagemodelsmostlyknow} study the calibration of prediction made by LLMs, and proposed two facets of model confidence: $P(T)$ and $P(IK)$.
$P(T)$, originally named as \textbf{P(True)}, represents the degree to which a model thinks the input statement is true.
For example, given a claim or statement, the value of $P(T)$ can be evaluated by prompting the model: \textit{``Is <statement> true? (A) True (B) False Answer: <answer>''}, and then measure the probability of the token ``\textit{(A)}'' predicted by the model at position \textit{<answer>}.

On the other hand, $P(IK)$ (stands for ``I Know'') describes how likely a model can give correct response to a query. The purpose of $P(IK)$ is to identify the problems that the model can or cannot answer, for the further goal of increasing the trustworthiness and reducing hallucination.
The evaluation of $P(IK)$ can be done in multiple ways: \citet{kadavath2022languagemodelsmostlyknow} trains a binary classification head on top of the language model by manually collecting training samples from various datasets. 
\citet{mahaut-etal-2024-factual} estimates $P(IK)$ with a similar setting to $P(T)$: prompting the model \textit{``Do you know the answer to
the following question: <question> (Yes/No/Maybe)? Answer: <answer>''} and measure the token probability of ``\textit{Yes}'' predicted by the model at position \textit{<answer>}.

When internal information of LLMs is not available, measuring the confidence of model responses becomes challenging.
One straightforward way is to use \textbf{\textit{verbalized confidence}} \cite{lin2022teachingmodelsexpressuncertainty}, which directly prompts the model to output confidence values for their response.
However, empirical results \cite{xiong2024llmsexpressuncertaintyempirical} have shown that LLMs tend to be overconfident while verbalizing their confidence level, leading to poor prediction performance.
Some black-box methods focus on prompt design, e.g., ask the model to provide multiple answers and their confidence at the same time, also known as \textit{top-k prompting} \cite{tian-etal-2023-just}.
Other methods investigate the distribution or semantic characteristics between multiple responses, such as \textit{\textbf{self-consistency}} \cite{wang2023selfconsistencyimproveschainthought} and \textit{\textbf{semantic entropy}} \cite{kuhn2023semanticuncertaintylinguisticinvariances}.
In sum, current black-box methods relies heavily on multiple sampling to learn the nuanced relationship between responses.
This would require significantly more computation at inference time compared to white-box methods, therefore they might not be the most efficient approach to estimate model confidence.

\subsection{Routing and Hybrid Inference}

Recent efforts in model routing and hybrid inference have introduced efficient strategies to balance LLM performance and cost.
Training-free methods like Eagle~\cite{zhao2024eagle} and RouteLLM~\cite{ong2024routellm} dynamically assign queries based on benchmark performance or preference-based scoring, while hybrid systems such as U-HLM architectures~\cite{oh2024uncertainty} combine small and large models for faster, cost-efficient inference. 
Hybrid LLM~\cite{dinghybrid}  also addresses this trade-off by training a query router that predicts response quality gaps between small and large models, enabling routing decisions tailored to the expected performance of each query.
Their approach estimates whether the small model can match the large model's response using BART scores, and tunes routing thresholds to control quality-cost tradeoffs.
SYMBOLIC-MOE~\cite{chen2025symbolic} further explores symbolic, skill-based routing to match inputs with appropriate experts without gradient-based optimization, combining model profiling with task-specific keyword extraction. 

Our work complements these by proposing a confidence-driven selection mechanism that integrates both token-level certainty ($P(T)$) and internal knowledge estimation ($P(IK)$) to decide when to escalate queries. Unlike prior routing models that often rely on offline training~\cite{shnitzer2024large}, symbolic rules, or task-specific heuristics, our approach enables dynamic, instance-level decision-making with minimal architectural overhead, and demonstrates robustness under both in-distribution and OOD conditions. 

\section{Models}
\label{sec:models}

The open-source models were obtained from HuggingFace with the following identifiers: 

\begin{itemize}[noitemsep]
  \item \texttt{meta-llama/Llama-3.2-3B-Instruct} (used for the 3B model)
  \item \texttt{meta-llama/Meta-Llama-3.1-8B-Instruct} (used for the 8B model)
  \item \texttt{meta-llama/Llama-3.3-70B-Instruct} (used for the 70B model)
  \item \texttt{Qwen/Qwen3-4B} (used for the 4B model)
  \item \texttt{Qwen/Qwen3-8B} (used for the 8B model)
  \item \texttt{Qwen/Qwen3-32B} (used for the 32B model)
\end{itemize}

All of the experiments were conducted on 4 NVIDIA RTX A6000 GPUs, each with 48GBs of VRAM.
For generation tasks, we used a deterministic decoding setup with \texttt{do\_sample=False}, and applied nucleus sampling with \texttt{top\_p=0.9} and \texttt{temperature=1.0}. 
These parameters were held consistent across all model configurations unless otherwise noted.
All results are based on a single run per model configuration without repeated sampling.

\section{Input Format}\label{sec:prompt}

The prompts used in this work are presented in Table~\ref{tab:prompt}.

\begin{table}[t]
    \centering
    \footnotesize
    \begin{tabular}{p{7.5cm}}
        \toprule
        \textbf{Prompt used for multiple choice questions.}\\
		The following are multiple choice questions about \{subject\}. Please determine the correct answer in the format ``Answer: \textless answer\textgreater'', where \textless answer\textgreater is A, B, C or D.\\

		Question: \{question\}\\
		Choices: \{choices\}\\
		Answer:\\

		\midrule
		\textbf{Prompt used for open-ended questions.}\\
		(system) The following are questions about entity relationship. Please determine the correct answer in the format ``Answer: \textless answer\textgreater''.\\
		\\
		(user) Question: \{question\}\\
		(assistant) Answer:\\

        \bottomrule
    \end{tabular}
    \caption{Prompt templates used in this work.}    \label{tab:prompt}
\end{table}

\section{Query Distribution Across  Stages}\label{sec:q_ct}

The number of queries processed at each stage is summarized in Table~\ref{tab:cc-query-counts}.
For example, in the 3B $\to$ 8B $\to$ 70B configuration, the compute cost is
$2 \times (487 \times 3 \times 10^9 + 198 \times 8 \times 10^9 + 745 \times 70 \times 10^9) = 1.1 \times 10^5$ GFLOPs.
In contrast, in the 8B $\to$ 70B configuration, the cost is
$2 \times (594 \times 8 \times 10^9 + 836 \times 70 \times 10^9) = 1.27 \times 10^5$ GFLOPs.
This illustrates that even though 3B$\to$8B$\to$70B involves more model stages, the total cost can be lower due to early-stage filtering, which reduces the number of queries reaching the larger models.

\begin{table}[t]
\centering
\small
\begin{tabular}{lccc}
\toprule
\textbf{LLM(s)} & \textbf{3B} & \textbf{8B} & \textbf{70B} \\
\midrule
3B $\to$ 8B & 487 & 943 & -- \\
3B $\to$ 70B & 487 & -- & 943 \\
8B $\to$ 70B & -- & 594 & 836 \\
\hline
3B $\to$ 8B $\to$ 70B & 487 & 198 & 745 \\
\bottomrule
\end{tabular}
\caption{Number of queries processed by each model stage under different configurations.}
\label{tab:cc-query-counts}
\end{table}

\section{Statistical Significance Analysis}
\label{sec:p_value}

To assess whether the observed differences in performance are statistically significant, we conducted McNemar's test on key model comparisons. 
The results are summarized in Table~\ref{tab:appendix-significance}.

The results indicate that in the cases of 8B $\to$ 70B vs. 70B and 70B $\to$ GPT-4o vs. GPT-4o, the p-values exceed 0.05. This suggests that there is no statistically significant difference in performance between these configurations. 
However, our method achieves comparable performance while significantly reducing computational costs.
These results support our claim that the proposed method offers an effective balance between performance and resource usage.

\begin{table}[t]
\centering
\small
    \begin{tabular}{lrr}
    \toprule
    \textbf{Comparison} & \textbf{Chi-squared} & \textbf{p-value} \\
    \midrule
    3B $\to$ 8B vs 8B & 4.00 & 0.0455 \\
    \textbf{8B $\to$ 70B vs. 70B}$^*$ & N/A & \textbf{0.4048} \\
    3B $\to$ 8B $\to$ 70B vs. 70B & 19.22 & 0.00001 \\
    3B $\to$ 70B vs. 70B & 20.48 & 0.00001 \\
    \hline
    3B $\to$ GPT-4o vs. GPT-4o & 15.84 & 0.00007 \\
    8B $\to$ GPT-4o vs. GPT-4o & 4.00 & 0.0455 \\
    \textbf{70B $\to$ GPT-4o vs. GPT-4o} & \textbf{0.22} & \textbf{0.6395} \\
    \bottomrule
    \end{tabular}
\caption{Significance test results. The exact p-value for 8B $\to$ 70B vs 70B is calculated using a binomial test instead of the chi-squared approximation.}
\label{tab:appendix-significance}
\end{table}

\section{Results with Open-ended QA}\label{sec:open-end}

One critical question for our framework is: \textit{"how can we generalize the method to generative tasks where multiple-choice $P(T)$ cannot be computed?"}
While some researchers have attempted to extend confidence estimation to generative tasks by either sampling multiple responses and transforming the original query into multiple-choice format \cite{ren2023selfevaluationimprovesselectivegeneration}, or by simply using average token probabilities over the generated output \cite{chuang2025confidentseekstrongerexploring}, the performance is modest.
In this paper, we propose a new confidence estimation technique called \textbf{\textit{first-token confidence}}.
We prompt the model to respond in a specific format and then measure the probability of the first token being \textit{``Answer''}.
The rationale behind this approach is that we observed models often deviate from the expected format when they are uncertain or unable to answer confidently.

To test our method on generative tasks, we perform evaluation using PopQA, a large-scale open-domain question answering dataset consisting of 14k entity-centric QA pairs.
A main problem we encountered while evaluating open-ended queries was the ineffectiveness of traditional keyword-based matching.
Previous work \cite{kamalloo-etal-2023-evaluating} emphasizes that exact match or simple overlap metrics understate LLM performance and miss hallucinations.
To mitigate this, we leveraged the grounding API to search for relevant results online and summarize whether the model's response can be considered correct.
Table~\ref{tab:accuracy_delta} shows the evaluation results with and without grounding.
Our results are in line with \cite{kamalloo-etal-2023-evaluating}, where simple lexical matching fails when models responded with a plausible answer.
This is especially the case when evaluating datasets like PopQA, whose questions are somewhat ambiguous and can have multiple correct answers.

\begin{table}[t]
\centering
\small
    \begin{tabular}{lcc}
    \toprule
    \textbf{} & \textbf{Accuracy} & $\Delta$ \\
    \hline
    3B (w/o grounding) & 0.1916 & - \\
    3B (w/ grounding)  & 0.2669 & +0.0753 \\
    \hline
    8B (w/o grounding) & 0.2520 & - \\
    8B (w/ grounding)  & 0.3514 & +0.0994 \\
    \hline
    70B (w/o grounding) & 0.4767 & - \\
    70B (w/ grounding)  & 0.6608 & +0.1841 \\
    \bottomrule
    \end{tabular}
\caption{Accuracy and delta for different model sizes with and without grounding.}
\label{tab:accuracy_delta}
\end{table}

For fair evaluation, we argue that a model responding \textit{"I don't know"} or asking for clarification should not be considered as incorrect.
We propose using \textbf{\textit{hallucination rate}} to reflect model's faithfulness, which evaluates how many responses are factually incorrect, verified by grounding API.

\begin{table}[t]
\centering
\scriptsize
\setlength\tabcolsep{1mm}
    \begin{tabular}{lccccc}
    \toprule
    \textbf{LLM(s)} & \textbf{Acc.} & \makecell{\textbf{Hallucination} \\ \textbf{Rate}} & \makecell{\textbf{Cost} \\ \textbf{(USD)}} & \textbf{PD} & \makecell{\textbf{Reduced} \\ \textbf{Cost}} \\
    \hline
    3B                & 0.2689 & 0.4169 & 0.0030 & - & - \\
    8B                & 0.3387 & 0.4679 & 0.0028 & - & - \\
    70B               & 0.6585 & 0.3208 & 0.0257 & - & - \\
    \hline
    3B $\to$ 8B         & 0.3317 & 0.4923 & 0.0027 & -2.07\% & 2.64\%  \\
    3B $\to$ 70B        & 0.6131 & 0.3547 & 0.0226 & -6.89\% & 12.29\%  \\
    8B $\to$ 70B        & \textbf{0.6459} & \textbf{0.3422} & \textbf{0.0239} & \textbf{-1.91\%} & \textbf{7.11\%}  \\
    \hline
    3B $\to$ 8B $\to$ 70B & 0.6075 & 0.3603 & 0.0219 & -7.74\% & 14.78\% \\
    \bottomrule
    \end{tabular}
\caption{Experimental results with LLaMA models on PopQA.}
\label{tab:accuracy_hallucination}
\end{table}

Table~\ref{tab:accuracy_hallucination} presents the accuracy, hallucination rate and cost of different LLaMA models on the PopQA dataset.\footnote{Table~\ref{tab:token-pricing} presents the token pricing of the models.}
As expected, larger models generally demonstrate higher accuracy and lower hallucination rates.
The 70B model achieves the best single-model performance, with an accuracy of 0.6585 and a hallucination rate of 0.3208, outperforming both the 3B and 8B variants by a significant margin.

When employing model cascades, we observe a trade-off between accuracy and hallucination. For example, the 3B $\to$ 8B $\to$ 70B cascade achieves an accuracy of 0.6075, which is slightly lower than using the 70B model alone.
The two-stage (8B $\to$ 70B) strategy achieves the highest accuracy (0.6459) and the lowest hallucination rate (0.3422) among all dynamic settings, while saving over 7\% in cost compared to using the 70B model exclusively.
Notably, the performance drop (PD) is minimal at only 1.91\%, demonstrating that dynamic model selection can yield promising results at a fraction of the cost.

\begin{table}[t]
\centering
\small
\setlength{\tabcolsep}{1mm}
    \begin{tabular}{lcc}
    \toprule
    \textbf{Model} & \textbf{Input (per 1M tokens)} & \textbf{Output (per 1M tokens)} \\
    \midrule
    3B & \$0.08 & \$0.16 \\
    8B & \$0.10 & \$0.20 \\
    70B & \$0.60 & \$1.20 \\
    GPT-4o & \$2.50 & \$10.00 \\
    \bottomrule
    \end{tabular}
\caption{Token pricing for different models.}
\label{tab:token-pricing}
\end{table}

\section{Cost Estimation}
\label{sec:cost}

To promote transparency and enable fair comparison, we provide a detailed cost analysis of all model configurations used in our experiments. 
While our initial assumption was that self-hosting LLMs would generally offer lower cost than API-based solutions, we conducted a systematic evaluation using publicly available token pricing information.

Specifically, we obtained pricing for input and output tokens from SambaNova's cloud-hosted LLaMA models (3B, 8B, and 70B),\footnote{\url{https://cloud.sambanova.ai/plans/pricing}} and compared them against OpenAI's GPT-4o API pricing.\footnote{\url{https://openai.com/api/pricing/}} 
We summarize the input and output token pricing of different models used in this work in Table~\ref{tab:token-pricing}.
All prices reported were recorded in April 2025.

For each setting, we calculated the total cost based on the number of input and output tokens consumed during inference.
Table~\ref{tab:cost-analysis} shows the per-token pricing as well as the estimated total cost across different model paths. 
As expected, direct use of GPT-4o incurs the highest cost (\$0.928 per run), while cascade-based configurations such as 70B$\to$GPT-4o (\$0.357) or 8B$\to$GPT-4o (\$0.650) offer a substantial reduction. 
Among open-source settings, 3B$\to$8B$\to$70B achieves strong performance while keeping the cost at only \$0.08, highlighting the method's efficiency in balancing accuracy and resource usage.
This analysis supports the utility of our approach in both open-source and API-based scenarios.

\begin{table}[t]
\centering
\footnotesize
    \begin{tabular}{lcc}
    \toprule
    \textbf{LLM(s)} & \textbf{Total Cost} & \textbf{Reduced Cost} \\
    \midrule
    3B & \$0.017 & - \\
    8B & \$0.021 & - \\
    70B & \$0.129 & - \\
    3B$\to$8B & \$0.020 & 4.76\% \\
    3B$\to$70B & \$0.096 & 25.58\% \\
    8B$\to$70B & \$0.089 & 31.01\% \\
    3B$\to$8B$\to$70B & \$0.080 & 37.98\% \\
    \hline
    GPT-4o & \$0.928 & - \\
    3B$\to$GPT-4o & \$0.707 & 23.81\% \\
    8B$\to$GPT-4o & \$0.650 & 29.96\% \\
    70B$\to$GPT-4o & \$0.357 & 61.53\% \\
    \bottomrule
    \end{tabular}
\caption{Estimated cost for each setting.}
\label{tab:cost-analysis}
\end{table}

\section{Evaluation of $P(IK)$}\label{sec:result_pik}
\begin{table}[t]
  \centering
  \small
    \begin{tabular}{lccc}
    \toprule
    \textbf{Model} & \textbf{Accuracy} & \textbf{F1} & \textbf{AUROC} \\
    \hline
    3B & 69.16\% & 62.76\% & 75.17\% \\
    8B & 76.50\% & 69.21\% & 81.78\% \\
    \bottomrule
    \end{tabular}%
  \caption{Performance of $P(IK)$ classifier of LLaMA 3B and 8B models.}
  \label{tab:p_ik_performance}
\end{table}%

Before evaluating the impact of our framework, we first examine the performance of the $P(IK)$ classifier itself. 
As shown in Table~\ref{tab:p_ik_performance}, the trained classifiers achieve decent performance, indicating that the $P(IK)$ classifiers can effectively estimate model's self-knowledge, providing a reliable signal for guiding model selection.
With this understanding, we can now analyze how incorporating $P(IK)$ affects the overall framework performance.

\section{Sensitivity Analysis of $P(T)$ threshold}
\label{sec:sensitivity}
We conducted a sensitivity analysis by varying the threshold of P(T), as shown in Table.
The threshold of $P(T)$ controls how confidently the smaller model must be before retaining the answer rather than routing it to the larger model.

In the 8B $\to$ 70B setting, we observe that as the $P(T)$ threshold decreases from 95\% to 50\%, the routing rate to the larger model increases (i.e., reduced compute cost improves), but this comes at the expense of a gradual decline in overall accuracy and performance drop (PD).
Importantly, with our full system (i.e., with both $P(T)$ and $P(IK)$), the accuracy remains relatively stable, dropping only ~0.8\% even when the threshold is lowered to 50\%, while reducing compute cost by \~39\%. However, when we disable $P(IK)$ and rely on $P(T)$ alone, the performance becomes more sensitive to the threshold. For instance, at a 50\% threshold, the accuracy drops significantly, despite higher compute savings.
These results confirm that $P(IK)$ complements $P(T)$ by stabilizing performance even when we allow more aggressive routing to the smaller model. It also suggests that thresholds around 90\% strike a good balance between accuracy and computation reduction.

\begin{table}[t]
\centering
\small
    \begin{tabular}{lrrr}
    \toprule
    \multicolumn{4}{c}{\textbf{8B $\to$ 70B}} \\
    \midrule
    \textbf{Setting} & \textbf{Accuracy} & \textbf{PD} & \textbf{Reduced CC} \\
    \midrule
    $P(T) \ge 95\%$ & 0.8350  & -0.08\% & 33.25\%  \\
    $P(T) \ge 90\%$ & 0.8322 & -0.42\% & 36.46\%  \\
    $P(T) \ge 70\%$ & 0.8287 & -0.84\% & 38.56\%  \\
    $P(T) \ge 50\%$ & 0.8266 & -1.09\% & 39.33\%  \\
    \midrule
    \multicolumn{4}{c}{\textbf{8B $\to$ 70B (w/o $P(IK)$)}} \\
    \midrule
    $P(T) \ge 95\%$ & 0.8350  & -0.08\% & 29.69\% \\
    $P(T) \ge 90\%$ & 0.8231 & -1.51\% & 38.15\% \\
    $P(T) \ge 70\%$ & 0.7923 & -5.19\% & 56.19\% \\
    $P(T) \ge 50\%$ & 0.7455 & -10.79\% & 74.24\% \\
    \bottomrule
    \end{tabular}
  \caption{Model Performance Comparison for LLaMA 8B $\to$ 70B.}
  \label{tab:model_comparison}
\end{table}

\section{GenAI Usage Disclosure}

We utilized generative AI tools solely for grammar and language refinement. 
All content was reviewed and edited by the author(s), who take full responsibility for the final manuscript.

\end{document}